# STRATEGY OF THE NEGATIVE SAMPLING FOR TRAINING RETRIEVAL-BASED DIALOGUE SYSTEMS

*Aigul Nugmanova[1,2], Andrei Smirnov[*], Galina Lavrentyeva[1,3], Irina Chernykh[1,3]*

[1]ITMO University, St. Petersburg, Russia
[2]Speech Technology Center, St. Petersburg, Russia
[3]STC-innovations Ltd., St. Petersburg, Russia
{nugmanova, lavrentyeva, chernykh-i}@speechpro.com, andrey.smirnov@juvo.ru

## ABSTRACT

The article describes the new approach for quality improvement of automated dialogue systems for customer support service. Analysis produced in the paper demonstrates the dependency of the quality of the retrieval-based dialogue system quality on the choice of negative responses. The proposed approach implies choosing the negative samples according to the distribution of responses in the train set. In this implementation the negative samples are randomly chosen from the original response distribution and from the "artificial" distribution of negative responses, such as uniform distribution or the distribution obtained by transformation of the original one. The results obtained for the implemented systems and reported in this paper confirm the significant improvement of automated dialogue systems quality in case of using the negative responses from transformed distribution.

## 1. INTRODUCTION

Automated dialogue systems in the customer support service recently become more popular area for research in the field of natural language processing[1]. One approach to develop such systems uses the retrieval-based dialogue models. Such models can use the unlabeled data during the training and their responses are predictable because they use only responses of the training set [2, 3, 4, 5]. For training these models it is important not only to customize the architecture, but also to create appropriate training data. For example, the most recent research [3] shows the impressive improvement of the quality for systems that used the weighting model for preparing training data [3].

In this paper we show how negative sampling strategy affects the performance of dialogue system. The main goal of the investigation the negative sampling methods are to form more effective training set. Random selection of negative samples allows to add a lot of identical examples to the train set if the original data contains repetitions. Our research shows how the train set can be drafted more diverse after simple transformations. Similar approach was described earlier in [6, 7, 8]. In [6] authors introduce negative sampling idea based on the concept of noise contrastive estimation (similar to generative adversarial networks), which implies, that a good model should differentiate fake signal. To achieve this goal several negative examples for every positive example are sampled from training data as noise examples and used to train the model. Authors use the noise distribution to choose negative samples by transformation the unigram distribution. We take it into account in our research and try to improve our systems by transforming the response distribution in order to choose more appropriate negative responses for retrieval-based dialogue systems training.

The dialogue systems investigated in this paper uses neural network architecture, performed in [4]. Neural network was used in two ways: for calculating the response probability for current question and for obtaining the text representation in order to find the nearest question.

Section 2 describes the architecture of the dialogue system. In section 3 the negative sampling is performed. Section 4 and section 5 contain the data description and definition of the evaluation metrics used during the experiments. Obtained results are reported in section 6. And section 7 concludes the produced investigation.

## 2. ARCHITECTURE

Dialogue systems, considered in this paper, are based on Siamese network like Dual Encoder Model, presented in [4].

---

[*] Work done while the author was at Speech Technology Center

It is the retrieval-based model. The main idea of this approach is to find the best response for current context The context here includes user's question and the previous utterances of the dialogue in training set.

We use this architecture in two ways and with two kinds of encoders. Fist approach uses the dual encoder model to find a pairwise probability of context and response like in the same approaches in [2, 4]. Second approach uses encoder model only to get sentences embeddings. In this case two types of neural encoder are considered: first is based on GRU cell and second uses Attention layer only.

### 2.1. Dual Encoder Model

Similar to [4] we use pair probability of context and response to find the best response.

The process of calculating the probability between current context and the response can be described as follows
- Context and response are divided into words sequences and initialized with the word embeddings. In this way two matrix with dimensions: sequence length, word embeddings dimension size are obtained
- These matrices are used as input layer of the encoder. As an output the encoder produces the representation for context and response sequences as illustrated in Figure 1
- For pairwise probability calculation the sigmoid function is applied to the product of context vector c with weight matrix M and response vector r.

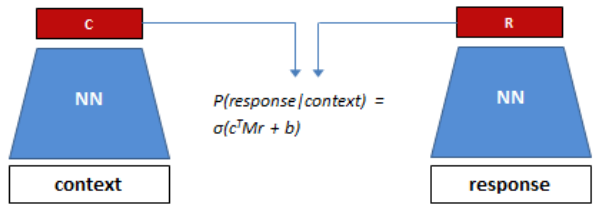

**Fig.1**. Scheme of Dual encoder model [2].

All the responses are then sorted by their probabilities. We presume that the highest-ranked response is the most appropriate response for current context.

The model used here is based on recurrent neural network with Gated Recurrent Unit (GRU) and hidden size 128. All models have the best result after 20000 iterations. For word embeddings we use N = 300 dimensional Word2Vec embeddings matrix pretrained on Russian dialogues corpus, included the target data and dialogues from popular web sites. Words of the training set which are not included in the pretrained model are initialized by average vector of word embeddings.

### 2.2. Embedding-Based Model

As an alternative approach, the architecture described in section 2.1 is used only for obtaining the contexts vectors. Here the output produced by the encoder is used to obtain the sentences representations.

In this approach we presume that the best response is in pair <nearest context, response> and we use this response as a correct answer. The similarity between current context vector and those from the training set is estimated using the cosine distance scoring. For searching the most similar context the representations of all context sequences in the training set are extracted. Further for current user question with previous dialog utterances, that all together contain the context of the dialog, the context representation vector is also obtained and the cosine distance is calculated.

In order to reduce the architecture of the network used for embedding extraction, the RNN layer was excluded from the original architecture leaving only the attention layer.

Our experiments show that the representations are more effective for employment n dialogue system if a linear combination of context and response representations is used for the search. It is called history vector and is expressed in (1).

$$history_i = context_i + c_r\, response_i \quad (1)$$

where $i$ is a number of pair in the training set, $context_i$ is a current context vector, $response_i$ is a current response vector, $c_r$ is a response weight. In our experiments we use $c_r$ as a free parameter and the best results are obtained for $c_r = 0.4$.

### 3. NEGATIVE SAMPLING

For training the systems, described in section 2, the negative sampling strategy is usually used. It helps to add the incorrect training examples into the training set. In this research we studied how the negative sampling strategy influences the quality of dialogue systems. We used several datasets prepared with the use of negative sampling methods described below.

For training the neural network with architecture described in Section 2 pairs <context, response> in each dialogue (where "context" is the concatenation of the current question and the previous utterances of the dialogue) are used as a training example of real (positive) responses. As negative samples N pairs <context, negative response> are used, where negative responses are incorrect answers selected from the training dialogues according to one of the techniques described below. We use a 1:5 ratio between positive and negative responses.

A popular approach to choose negative responses for concrete context is a random response selection from others dialogs. We suppose this approach is not optimal, because the most uninformative frequent utterances fall into subsamples more often than rare informative utterances. To overcome this problem we suggest to change the responses distribution and to choose responses for negative samples from the transformed distribution.

In our experiments 4 methods of negative response selection are considered:

- The real response distribution is taken into account. Responses for negative samples are selected randomly.
- The response distribution is transformed to the uniform distribution and responses for negative samples are selected from the obtained one.
- The response distribution is transformed to the distribution obtained by raising the initial distribution to some power and responses are selected from the new distribution. It is important to note that negative degree helps to reduce the amount in frequent sentences of the base among negative samples.
- The latter approach also aims at bringing the distribution closer to the uniform. But in this method, the responses distribution influences not only the choice of the negative answer, but also the probability of the example entering the training set. For this purpose, the amount of occurrences in the dataset of dialogues for each answer from the current pair <context, answer> is calculated (N). The pair <context, answer> is added to the set of training data only with probability 1 / N.

To take into account the semantic similarity between phrases and to approximate the probability density for responses in the dataset by a continuous density function we apply a kernel-density estimation using Gaussian kernels. In our experiments we use the bandwidth value 0.4.

## 4. DATA

In this paper for training and evaluation the proposed method the Russian language dataset with human-human unstructured conversation without any labels was used. The dataset is a chat log of technical support of the web portal. It contains 25000 dialogues with average length of 4 turns.

**Table 1.** Example of dialogue between a user and an operator.

| English (translate): |
| --- |
| **Q1:** Hello! How can I register in the web service? **A1:** Hello, my name is <name> and I will be glad to help you. Registration on the portal is available by a personal visit to the Service Center or on your own, which of the following ways would be more comfortable for you? |
| **Q2:** Thanks for the answer, I will come myself. **A2:** For registration, you need to contact the Service Center that is convenient for you. You can see the addresses by clicking on the <link>. You need to have a passport and SNILS. |
| **Q3:** OK. **A3:** Do you have any question about the portal? |

Table 1 demonstrates conversation examples translated to English language. The data were divided with ratio 80:10:10 for training, automatic evaluation and human evaluation.

Dialogues presented in the dataset contain a big amount of uninformative utterances. In this case the amount of uninformative responses in the training set will be much bigger than the amount of informative ones, which will lead to low performance of the final model. For example, the beginning of the most dialogues includes greetings and the ending of dialogues includes valedictions. Sometimes an operator can ask users to wait while he is looking for the information. Also, the examples of frequently responses are "Yes" or "No".

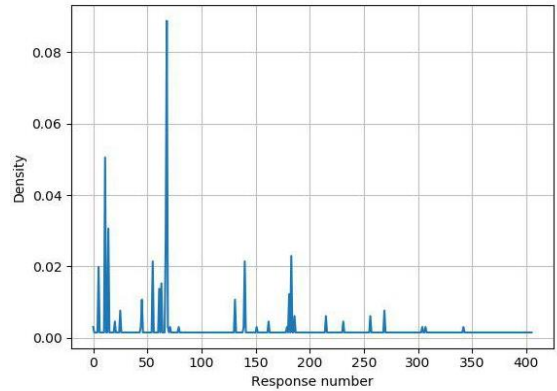

**Fig.2.** The response distribution.

Figure 2 illustrates the responses distribution curve for the first 1000 dialogues in the dataset. This curve demonstrates that the set contains phrases with very high frequency. Most of them are uninformative. Reducing the influence of these responses is one of the aims to investigate negative sampling methods.

## 5. EVALUATION METRIC

To evaluate the quality of the proposed models the automatic and human evaluation methods are used. Both are based on the recall@k metric, similar to the evaluation of the retrieval-based systems in [2, 3, 4].

### 5.1. Automatic evaluation

Test set includes 2500 dialogues. For each pair <context, response> sampled from the test set m alternative negative responses are selected. These m+1 responses are then ranked according to its similarity to the context. The output is defined as the right answer if the original <context, response> pair appears in top-k among all m+1 candidates. In experiments we use m=9.

### 5.2. Human evaluation

We also apply a human evaluation in our research. For human evaluation 400 test questions were specially selected by the experts from the corresponding test set. These questions contained only targeted questions requiring a meaningful response. In the experiments the responses are chosen according to the ranking training dialogues and select from the responses with the highest probabilities. The responses obtained by several models trained on the data with negative sampling from different distribution are evaluated.

**Table 2:** Recall@3 for correct responses (CR) and unsure responses (UR) based on human evaluation for models trained using the data with different distributions in negative sampling (NS).

| Approaches | | randomly NS (baseline) | NS from uniform distribution | NS from base distribution in -0.125 degree | NS from base distribution in -0.25 degree | NS from uniform distribution and filtered dialogues |
|---|---|---|---|---|---|---|
| **DE GRU** | UR | 0.40 | 0.45 | 0.49 | 0.46 | 0.44 |
| | CR | 0.24 | 0.23 | 0.22 | 0.22 | 0.20 |
| **DE emb GRU** | UR | 0.76 | 0.79 | 0.77 | 0.75 | 0.8 |
| | CR | 0.42 | 0.45 | 0.46 | 0.45 | 0.46 |
| **DE emb ATT** | UR | 0.7 | 0.7 | 0.71 | 0.72 | 0.71 |
| | CR | 0.40 | 0.44 | 0.41 | 0.47 | 0.48 |

Our model selects three responses for each test question. For each of 1200 selected responses two assessors rate the consistency between context and response using a 4-points scale. The response is marked as: 0, if the response is incorrect; 1, if the response can be interpreted as correct by the user which is not an expert in the field; 2, if the response include information of correct answer; 3, when it is a reference answer.

Also we take into account that the human marks can be changed over time between evaluations and therefore we fill in the test table by responses of different models and then shuffle it.

Based on the results of estimates two metrics are calculated: recall@3 for correct response (CR) and recall@3 for unsure response (UR). Recall@3 for correct response is equal 1 if in three responses selected by the model there is at least one with human mark above 1. Similarly, recall@3 for unsure response is declared to be 1 if in three responses there is at least one with human mark above 0.

## 6. RESULTS

At first, we tested our models automatically with recall@k metrics. In the test set the alternative responses was chosen from the distribution of the training set. The model was trained on the training set with negative samples from initial and uniform distributions. Each model was then evaluated on the test sets with alternative responses from both of these distributions.

The results presented in table 3 confirm that models show poor quality on the test samples with transformed response distribution. This indicates that for automatic evaluation it is important that test and training responses are sampled from the same distribution. Otherwise the actual increase of the dialogue system quality with different negative sampling strategies cannot be estimated.

We presume that the human evaluations show the difference between models better. Table 2 presents the CR and UR (such as in Section 5.2) metrics for three models: Dual Encoder with GRU cell (DE GRU), embeddings from Dual Encoder with GRU cell (DE emb GRU) and embeddings from Dual Encoder with Attention layer (DE emb ATT).

Evidently, any changes in the response distribution, which align it, leads to higher quality of dialogue systems based on embedding from encoder in terms of human marks. Moreover when we use the dual encoder model to rank responses in the training set, we can use the degree of the response distribution as a free parameter and achieve improvement by selecting the more suitable degree value. For example we achieved the best quality using the degree=-0.125. Also table 2 demonstrates that filtering dialogues during the training can be effective for text representations, but it does not improve the dual encoder based model.

Comparing models experiments show that when the training set is not big the embeddings-based model works much better than the full model of the dual encoder (up to 2 times in our case with 20000 dialogues). Also in Table 2 it is noticeable that the model using the GRU in the encoder shows the result higher than the analogous architecture with only attention layer.

**Table 3:** Recall@1 values for GRU dual encoder model with negative responses from the original response distribution and uniform distribution.

| Test set (alternative responses) | Training set (negative samples) | Recall@1 |
|---|---|---|
| **initial** distribution | **initial** | **0.57** |
| | uniform | 0.45 |
| transformed distribution (**uniform**) | initial | 0.61 |
| | **uniform** | **0.69** |

## 7. CONCLUSION

This paper reports the detailed analysis of negative sampling strategy for training retrieval-based dialogue systems with several architectures. The conducted experiments confirm that using the proposed negative sampling strategy instead of the random sampling, helps to achieve a relative improvement up to 20% in terms of the dialogue system quality by human evaluation. It is also shown that the embedding based model demonstrates twice better results than the full dual encoder model on our data.

Our future objective is to consider other methods of negative sampling with the use of additional information extracted from the data such as topic clustering or number of turns in dialogues.